\documentclass{article}

\usepackage[main, final]{neurips_2026}

\usepackage[utf8]{inputenc} 
\usepackage[T1]{fontenc}    
\usepackage{hyperref}       
\usepackage{url}    
\usepackage{caption}
\usepackage{booktabs}       
\usepackage{amsfonts}       
\usepackage{nicefrac}       
\usepackage{microtype}      
\usepackage{xcolor}         
\usepackage{algorithm}
\usepackage{algorithmic}
\usepackage{amsmath} 
\usepackage{makecell}
\usepackage{multirow}
\usepackage{graphicx}
\usepackage{wrapfig} 
\usepackage{arydshln}
\usepackage{cleveref}
\usepackage{duckuments}
\usepackage{enumitem}

\newcommand{\Mix}{\textsc{Mix}}
\newcommand{\SFT}{\textsc{SFT}}
\newcommand{\Merge}{\textsc{Merge}}

\newcommand{\std}[1]{{\scalebox{0.55}{$\pm$#1}}}
\definecolor{citationcolor}{RGB}{80, 90, 180}
 \hypersetup{
     colorlinks=true,
     linkcolor=citationcolor,
     filecolor=blue,
     citecolor=citationcolor,      
     urlcolor=cyan,
     }

\title{Mixture-Trained Merging for \\ Unified Multi-Objective Models}

\author{%
  SeongHyeon Kim$^{1}$\thanks{Equal contribution.} \quad
  Chaeyun Jang$^{1}$\footnotemark[1] \quad
  Seungyoo Lee$^{1}$ \quad
  Jiyeon Ham$^{2}$ \quad
  Yunju Bak$^{2}$ \\
  \textbf{Boseop Kim}$^{2}$ \quad
  \textbf{Juho Lee}$^{1}$\thanks{Corresponding author.} \\[0.5ex]
  $^{1}$KAIST \quad $^{2}$Kakao \\[0.3ex]
  \texttt{\{shkim0824, jcy9911, punctuate, juholee\}@kaist.ac.kr} \\
  \texttt{\{chloe.py, juliet.bak, mat.mul\}@kakaocorp.com} \\
}

\begin{document}

\maketitle

\begin{abstract}
Unified language models are increasingly expected to combine heterogeneous
capabilities, such as mathematics, code, instruction following, and controllable
thinking behavior, within a single set of parameters. A common solution is
sequential post-training on multiple objectives, but this entangles all
objectives along one optimization trajectory and makes the final model highly
sensitive to training order, data ratios, schedules, and stopping criteria.
Weight-space merging offers a modular alternative, but naive merging of
single-objective experts often fails: domain capabilities degrade sharply, or
\texttt{think}/\texttt{non-think} modes collapse into one dominant behavior. We
attribute both failures to incompatible weight-space geometry: experts trained
on single objectives drift to distant regions of parameter space, placing
their interpolations outside any shared low-loss basin. We propose \textbf{Mixture-Trained Merging (MTM)}, which trains each branch on
an objective-biased data mixture rather than a single objective, exposing it to
cross-objective interactions and making branches compatible at merge time. MTM
uses merged-model evaluations as a low-cost signal for selecting branch
mixtures, avoiding expensive data-mixture ablations. The procedure is iterative:
each round promotes the base model using globally selected merge coefficients
and refines each branch mixture using domain-preferred coefficients under
constraints that preserve other objectives. To scale beyond simplex grid search,
MTM uses qNEHVI-based multi-objective Bayesian optimization. Across code,
mathematics, instruction following, and \texttt{think}/\texttt{non-think}
control, MTM outperforms naive merging and preserves behavioral separation where
single-objective merging collapses, suggesting that effective unified models
require branches trained to be mergeable.
\end{abstract}

\section{Introduction}
Recent large language models (LLMs) increasingly aim to unify diverse capabilities within a single model, including complex reasoning, direct response generation, coding, and instruction following~\citep{guo2025deepseek, yang2025qwen3, wang2025nemotron, yang2025r}. Beyond performance, such unified models are practically appealing because they reduce the need for separate model stacks and offer a simpler deployment interface across heterogeneous workloads. 

A common practical approach to building such unified models is single-trajectory training, in which a model is post-trained across multiple objectives within a shared set of parameters~\citep{yang2025qwen3, yang2025r, wang2025nemotron}. While conceptually simple, this strategy introduces substantial engineering overhead. The final performance is often highly sensitive to training order, data mixing ratios, and scheduling choices, because these factors strongly affect objective interference and catastrophic forgetting~\citep{wang2025see, luo2025empirical, koh2026msft}. As a result, finding a robust recipe typically requires extensive hyperparameter tuning over near-full-budget runs, making unified model construction costly and difficult to iterate on.

Weight-space merging, which trains specialized branches separately and then combines them, offers an appealing alternative to optimizing all objectives along a single training trajectory~\citep{li2022branch, yadav2023ties, ilharco2023editing}. This makes unified model construction more modular and avoids much of the optimization complexity of single-trajectory training.
Merging aims to combine specialized branches into a single model that performs well on all target objectives. However, we empirically find that merging single-objective experts does not always lead to such a model.
For example, when constructing a hybrid model that should support both non-think and think modes, merging a non-think expert with a think expert collapses the two modes toward the branch with the larger merge coefficient, producing uniformly verbose or concise outputs regardless of the requested format (\Cref{fig:think_merge}). Similarly, when constructing a multi-domain model that should handle both code and math, merging a code expert with a math expert leads to broad downstream degradation across both domain accuracy, as shown in \Cref{fig:domain_merge}.

We hypothesize that these failures arise because independently trained branches drift into different regions of weight space. Consistent with this view, we observe substantial accuracy drops along the linear interpolation path between high-performing branches, indicating that they do not lie on an accuracy-preserving path~\citep{garipov2018loss, izmailov2018averaging}. This suggests that mergeability must be encouraged during training rather than imposed only through post-hoc merging. We therefore train each branch on a controlled \textbf{mixture} of objectives. By exposing branches to non-primary objectives, mixture training reduces inter-branch divergence while preserving specialization. Unlike single-trajectory multi-objective training, which searches for a single global data ratio, our setting requires branch-specific mixtures that yield specialized yet mutually mergeable branches.

To identify these mixtures, we use the connection between weight-space merging and data-mixture training established by \cite{wang2026mergemix}: merge coefficients can approximate data-mixture ratios under first-order training dynamics. We use merged performance as a proxy signal and optimize per-branch mixtures with Multi-Objective Bayesian Optimization (MOBO)~\citep{brochu2010tutorial, frazier2018bayesian}.
Rather than running mixture search as a separate stage, we integrate it into an iterative branch specialization and merging procedure. Each round selects mixtures, continues branch training, and merges branches back into a common reference model to control divergence. Thus, mixture selection, specialization, and realignment are coupled within a single training loop.

Our contributions are summarized as follows:
\begin{itemize}
    \item We show that merging hard single-objective experts induces task interference, manifesting as mode collapse in think/non-think merges and broad degradation in domain-level merges.
    \item We connect this failure to weight-space incompatibility, showing that independently trained branches exhibit accuracy barriers along linear interpolation paths.
    \item We introduce iterative branch specialization and merging, which optimizes branch-specific objective mixtures to preserve both specialization and mergeability.
    \item We demonstrate consistent gains on Qwen3-4B and OLMo-7B across diverse math, code, and instruction-following benchmarks.
\end{itemize}

\section{Related Works}
\label{sec:related_works}
\subsection{Building Unified Models via Data Mixtures}
Modern LLMs are expected to handle multiple objectives such as math,
code, and instruction following within a single
model~\citep{yang2025qwen3, olmo2025olmo}, which makes the choice of
the SFT data mixture a central design decision. A growing line of
work therefore aims to determine mixture ratios in a principled
way. DoReMi~\citep{xie2023doremi} reweights domains via a small
proxy model trained with group distributionally robust optimization.
RegMix~\citep{liu2024regmix} fits a regression model from many
small-scale runs to predict performance of unseen mixtures. Data
mixing laws~\citep{ye2024data} formulate scaling laws for the
mapping from mixture proportions to validation loss. MergeMix~\citep{wang2026mergemix}
bypasses surrogate training entirely by approximating mixture-trained
models with linear weight-space merges of objective-specific
fine-tunes. Even with these advances, identifying a single global
mixture that simultaneously balances all objectives still relies on
heuristic choices and remains difficult to tune as the set of
objectives grows.

\subsection{Weight-Space Model Merging}

Model merging offers a parameter-space alternative to multi-objective
training, combining multiple specific objective models into a single model
without increasing inference cost~\citep{wortsman2022model,
ilharco2023editing, yadav2023ties, yu2024language, wan2025fusechat}.
Model Soups~\citep{wortsman2022model} showed that averaging weights
from multiple fine-tuned checkpoints can outperform the best individual
checkpoint, while task arithmetic~\citep{ilharco2023editing} generalized
this idea by representing fine-tuned models as task vectors and
combining them through linear operations in weight space. For LLMs,
Branch-Train-Merge~\citep{li2022branch} trains domain experts from a
shared branching point to keep them close enough in parameter space to
remain mergeable.

A central challenge in merging is parameter interference between objectives. Several methods address this interference post hoc: TIES
Merging~\citep{yadav2023ties} trims low-magnitude updates and resolves
sign conflicts, DARE~\citep{yu2024language} sparsifies and rescales SFT
deltas as a plug-in for existing mergers, and SCE~\citep{wan2025fusechat}
sets fine-grained fusion coefficients from update magnitudes. More
recently, application-specific mergers such as
RAIN-Merging~\citep{huangrain} target the fusion of reasoning and
instruction-tuned models by constraining thinking-format-aware
subspaces. In contrast to these post-hoc correction methods, MTM does
not take independently trained experts as fixed. Instead, it trains each
branch on a domain-biased mixture, shaping task vectors to be mutually
compatible before merging.
\begin{figure}[t]
  \centering

  \begin{minipage}[t]{0.48\textwidth}
    \centering
    \includegraphics[width=\linewidth]{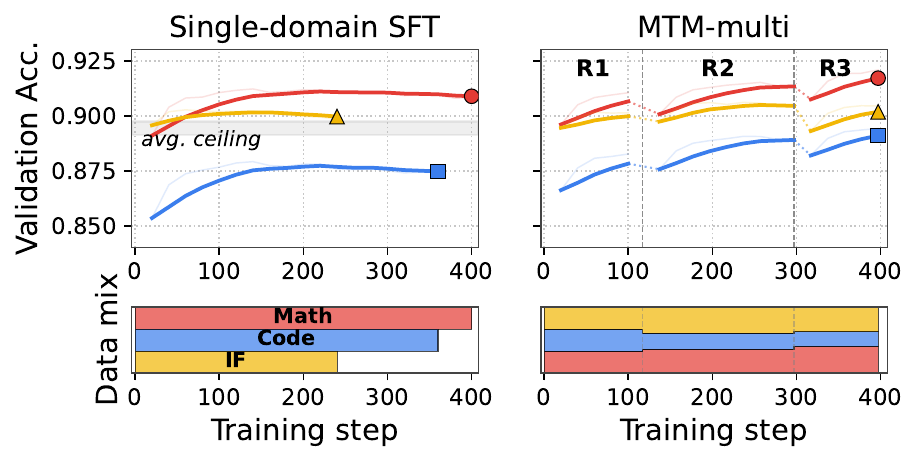}
    \caption{\textbf{Training dynamics of single-domain SFT vs. MTM-Multi in Qwen3-4B-Base.} 
    MTM-Multi replaces one long heterogeneous trajectory with round-wise
    mixture-trained branches and promoted merges.}
    \label{fig:training_steps}
  \end{minipage}
  \hfill
  \begin{minipage}[t]{0.48\textwidth}
    \centering
    \includegraphics[width=\linewidth]{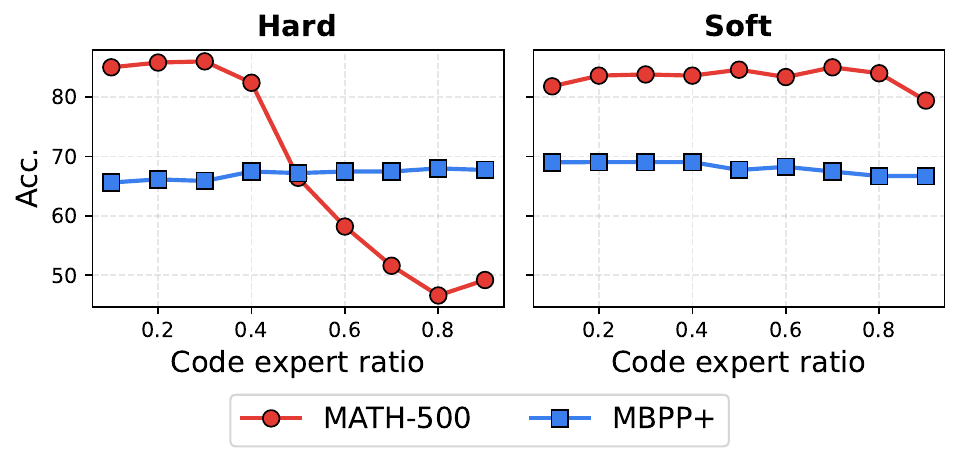}
    \caption{\textbf{Domain-expert merging on Qwen3-4B-Base.} Following \Cref{sec:experiments}, \textbf{Hard} branches are trained on single domains, whereas \textbf{Soft} branches use complementary Math/Code mixtures with a $(0.7,0.3)$ split.}
    \label{fig:domain_merge}
  \end{minipage}
\end{figure}

\begin{figure}
  \centering
  \includegraphics[width=\textwidth]{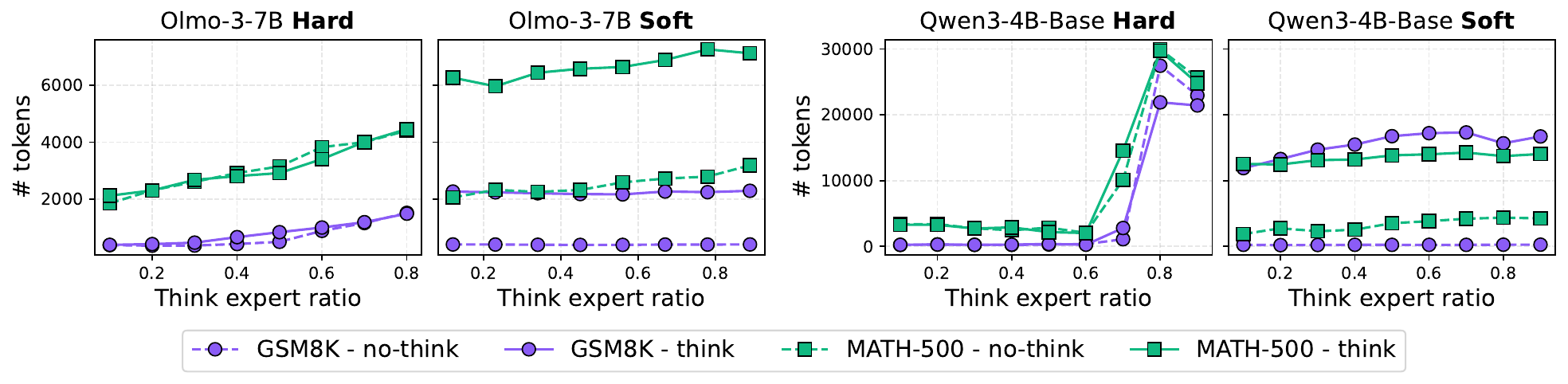}
    \caption{\textbf{Mode-expert merging on OLMo-7B and Qwen3-4B-Base.} Following \Cref{sec:experiments}, \textbf{Hard} branches are trained on single response modes, while \textbf{Soft} branches use complementary \texttt{think/non-think} mixtures with a $(0.7,0.3)$ split. Merging hard experts loses mode control under large branch imbalance, leading to outputs that follow the dominant branch (think or non-think) regardless of the prompt, whereas mixture-trained branches better preserve mode distinction.}
  \label{fig:think_merge}
\end{figure}

\section{Background and Motivation}
\label{sec:background}
\paragraph{Problem Setting}
We start from a base model $\mathcal{M}_0=\mathcal{M}(\theta_0)$ and $K$
objectives, each with a training corpus $\mathcal{D}_k$ and an evaluation
metric $f_k$ normalized to a common scale. We seek a single parameter
vector $\theta$ such that $\mathcal{M}(\theta)$ performs well on all
$\{f_k\}_{k=1}^{K}$. Throughout, $\Delta^{K-1}$ denotes the $(K{-}1)$-simplex,
$\Mix(\mathcal{D}_{1:K};q)$ the mixture corpus that draws from $\mathcal{D}_k$
with probability $q$, and $\SFT(\mathcal{M},\mathcal{D})$ supervised
fine-tuning of $\mathcal{M}$ on $\mathcal{D}$. Given $K$ models
$\{\mathcal{M}_k\}_{k=1}^{K}$ fine-tuned from a common reference
$\mathcal{M}_{\mathrm{ref}}=\mathcal{M}(\theta_{\mathrm{ref}})$ and
coefficients $\alpha\in\Delta^{K-1}$, we write
\begin{equation}
\Merge(\mathcal{M}_{1:K};\alpha)
\;=\;
\mathcal{M}\!\left(
\theta_{\mathrm{ref}}
+
\sum_{k=1}^{K}\alpha_k(\theta_k-\theta_{\mathrm{ref}})
\right)
\end{equation}
for the linear merge.

A standard approach to building such a multi-objective model is to train a
single model on an appropriately chosen global data mixture.
However, \Cref{fig:training_steps} shows that single-domain SFT runs on
math, code, and instruction following saturate at different training
steps and at different metric values. A single-trajectory recipe must
therefore compress this heterogeneity into one training schedule. This
makes recipe design a high-dimensional engineering problem in which the
mixture ratio, stage ordering, learning-rate schedule, and stopping
criterion must all be chosen so that no objective is undertrained while
others have already saturated, despite their intrinsically mismatched
saturation profiles.

A natural alternative is to train objective-specific branches in parallel
and combine them in weight space. For each $k$, we fine-tune a hard expert
$E_k = \SFT(\mathcal{M}_0,\mathcal{D}_k)$ and merge with weights
$\alpha\in\Delta^{K-1}$ as
$\mathcal{M}_{\textsc{merge}}(\alpha) = \Merge(E_{1:K};\alpha)$.
This decouples branches and reduces unified-model construction to many
cheap, independent fine-tunes plus a simple mixing step. Most existing
weight-space merging methods~\citep{ilharco2023editing, li2022branch} operate in this regime: take hard experts as given, then post-hoc combine them.

\subsection{Failure Modes of Hard-Expert Merging}
\label{sec:failure_modes}
In this section we
examine two failure modes that arise across our experimental settings.
First, hard-expert merging exhibits a sharp trade-off between target
domains: gains on one domain come at the cost of another, so no merge
ratio recovers all per-domain experts at once. Second, when merging a
non-think expert with a think expert to obtain a hybrid model, the
resulting model collapses into a single mode rather than retaining
both behaviors as intended. Both failure modes are alleviated when
the branches are trained on objective mixtures rather than as hard
experts, motivating the approach we develop in \Cref{sec:method}. 

\paragraph{Specialization at the cost of other domains.}
Hard experts specialize in their target domain, but underperform the
base model in other domains. As a result, their linear merge recovers
expert-level performance only near the simplex endpoints, as shown in
the left panel of \Cref{fig:domain_merge}. As the code-expert ratio
increases, MBPP+ accuracy improves, but MATH-500 accuracy drops
sharply from about $85$ to below $50$. Thus, no interior merge ratio
reaches expert-level performance on both objectives. In contrast, the
right panel shows that mixture-trained branches maintain strong
performance on both tasks across a wide range of merge ratios,
indicating that mixture-trained components are easier to combine.

\paragraph{Mode collapse between qualitatively different branches.}
The problem becomes more pronounced when branches encode different
generation behaviors. A non-thinking expert is trained to give short
direct answers, whereas a thinking expert produces longer
Chain-of-Thought (CoT) reasoning. A useful hybrid model should
preserve both behaviors and switch between them according to an
explicit mode signal, but linear interpolation of hard experts fails
to do so. In the left panels of \Cref{fig:think_merge}, generation
lengths under \texttt{no-think} and \texttt{think} prompts move
together as the think-expert weight changes, so the mode gap largely
disappears across the interior of the simplex; Qwen3-4B-Base
additionally shows a length blow-up at one merge ratio. In contrast,
the right panels show that mixture-trained branches preserve the
separation across merge ratios, maintaining short \texttt{no-think}
outputs and long \texttt{think} outputs simultaneously.

\section{MTM: Mixture-Trained Merging}
\label{sec:method}

We now present Mixture-Trained Merging (MTM), which addresses the
failures of hard-expert merging identified in
Section~\ref{sec:background} by training each branch on a mixture of
objectives. MTM has two variants. The
single-round variant (Section~\ref{sec:mtm:search}), \textbf{MTM-Single},
trains mixture-biased branches once and selects merge weights on top of
them using the MergeMix approximation. The iterative variant
(Section~\ref{sec:mtm:iterative}), \textbf{MTM-Multi}, spreads the same
budget across several shorter rounds and reuses each round's merged
model as the starting point of the next, so that mixture search and
model construction share a single training trajectory. The candidate
merge weights evaluated in either variant are produced by multi-objective
Bayesian optimization over the simplex (Section~\ref{sec:mtm:bo}).


\subsection{Merge-Based Mixture Search}
\label{sec:mtm:search}

\Cref{sec:background} showed that merging hard experts fails because the experts drift into mutually incompatible regions of weight space. MTM addresses this by replacing each hard expert with a \emph{soft-expert}: a branch trained on a domain-biased
\emph{mixture} of objectives. Each soft-expert remains specialized
toward its target domain keeping it close enough to the other branches in weight space. The unified model is then constructed by merging these $K$ soft-experts.
We initialize the reference model as the base model,
$\mathcal{M}_0=\mathcal{M}(\theta_0)$, and associate each branch
$k\in[K]$ with its own mixture $p_k\in\Delta^{K-1}$ that is biased
toward domain $k$. Each soft-expert is trained on its mixture,
\begin{equation}
E_k
=
\SFT\!\left(
\mathcal{M}_0,
\Mix(\mathcal{D}_{1:K};p_k)
\right),
\label{eq:branch-train}
\end{equation}
and the $K$ soft-experts are combined with merge weights
$\alpha\in\Delta^{K-1}$ to form the final unified model,
\begin{equation}
\mathcal{M}(\alpha)
=
\Merge(E_{1:K};\alpha),
\qquad
F^{(j)}(\alpha)
=
f_j\!\left(\mathcal{M}(\alpha)\right),
\label{eq:round-merge}
\end{equation}
where $F^{(j)}(\alpha)$ is the score of the merged model on objective
$j$. Crucially, each $p_k$ is a \emph{branch-specific} mixture: branch
$k$ leans toward $\mathcal{D}_k$ but contains controlled amounts of
the other objectives, so that the $K$ branches together cover the
simplex of objectives from $K$ different angles rather than collapsing
onto a single shared mixture.

This raises the question of how to choose the $K$ branch mixtures
$\{p_k\}_{k=1}^{K}$ without fine-tuning a new model for every
candidate. For this \emph{search} step, we borrow the approximation of
\cite{wang2026mergemix}: in a local fine-tuning regime, the behavior
of a linear merge $\Merge(E_{1:K};\alpha)$ approximates that of a model
trained on the corresponding data mixture
$\Mix(\mathcal{D}_{1:K};\alpha)$. We can therefore evaluate candidate
mixture ratios by evaluating merge weights on already-trained branches,
turning mixture search into an evaluation problem over the merge
simplex. Whereas MergeMix uses this approximation to pick a single
global mixture and trains one model on it, MTM uses it to pick $K$
branch-specific mixtures and deploys their merge. We refer to the
resulting single-round procedure as \textbf{MTM-Single}.

\subsection{Iterative Merging with Branch Reuse}
\label{sec:mtm:iterative}

MTM-Single decouples mixture search from training, but at a cost: the
branches trained for search and the model deployed at the end are
different objects, so producing a final unified model still requires a
second, full training pass on the chosen mixture. The total budget is
roughly twice that of a single-trajectory baseline, and the final
model never benefits from the merge evaluations carried out during
search.

A more efficient alternative is to spread the same total budget over
several shorter rounds, and at each round use the merge-based search
of Section~\ref{sec:mtm:search} not only to inform the next mixture
but also to construct the model itself. The merged model selected at
round $t$ becomes the reference model from which round $t{+}1$
branches are trained, so the search trajectory and the training
trajectory are the same sequence of models. We call this iterative
variant \textbf{MTM-Multi}. Each round $t$ performs three steps:
\begin{enumerate}[leftmargin=1.5em,itemsep=2pt,topsep=2pt]
\item train branches $E_{1:K}^{(t)}$ from the current reference model
$\mathcal{M}_{t-1}$ on mixtures $\{p_k^{(t)}\}$;
\item evaluate a set of candidate merges over the simplex to obtain a
score landscape $\{F_t^{(j)}(\alpha)\}$;
\item use this landscape to choose both the next reference model
$\mathcal{M}_t$ and the next-round mixtures $\{p_k^{(t+1)}\}$.
\end{enumerate}
Reusing the round-$t$ branches in step (iii) -- rather than discarding
them after search -- is what brings the total training cost back in
line with a single-trajectory baseline while preserving branch-level
modularity. The remainder of this subsection details steps~(i)--(iii).

\paragraph{Training mixture-trained branches.}
We initialize the first-round mixtures at the simplex vertices,
$p_k^{(1)}=e_k$ for $k\in[K]$, so round~1 starts from hard-domain
branches.
Round-$t$ branches are obtained by applying
Eq.~\eqref{eq:branch-train} from the current reference model
$\mathcal{M}_{t-1}$ with branch mixtures $\{p_k^{(t)}\}$. Round 1 thus
trains pure-domain experts from $\mathcal{M}_0$, and later rounds
train mixture-biased branches from a merged starting point.

\paragraph{Promoting the reference model.}
After evaluating a candidate set $\mathcal{X}_t\subset\Delta^{K-1}$,
whose construction is deferred to Section~\ref{sec:mtm:bo}, we select
a global merge weight that maximizes mean normalized performance,
\begin{equation}
    \alpha_\star^{(t)} \;=\;
    \arg\max_{\alpha\in\mathcal{X}_t}\;
    \frac{1}{K}\sum_{j=1}^{K} F_t^{(j)}(\alpha),
    \label{eq:global-ratio}
\end{equation}
and promote the corresponding merged model to the next-round reference,
\begin{equation}
    \mathcal{M}_t \;=\; \Merge\!\left(E_{1:K}^{(t)};\, \alpha_\star^{(t)}\right).
    \label{eq:promote-base}
\end{equation}
This step plays a dual role. It produces the unified model that MTM
ultimately returns at round $T$, and it acts as a \emph{realignment}
operation between rounds: collapsing the branches back into a common
reference prevents the cumulative drift that otherwise breaks
weight-space compatibility over long fine-tuning horizons.

\paragraph{Updating branch mixtures.}
For each branch $k$, we look for a merge weight that produces a strong
domain-$k$ specialist \emph{without} sacrificing the other objectives,
\begin{equation}
    \alpha_k^{(t)} \;=\;
    \arg\max_{\alpha\in\mathcal{X}_t}\; F_t^{(k)}(\alpha)
    \quad\text{s.t.}\quad
    F_t^{(j)}(\alpha)\;\geq\;\tau_j^{(t)},
    \quad \forall j\neq k,
    \label{eq:specialist-ratio}
\end{equation}
where $\tau_j^{(t)}$ is a per-objective preservation threshold expressed
on normalized scores, e.g., the score of $\mathcal{M}_{t-1}$ on
objective $j$. The selected $\alpha_k^{(t)}$ then propagates back to
the data mixture used to train branch $k$ in the next round,
\begin{equation}
    p_k^{(t+1)} \;=\; \sum_{\ell=1}^{K} \alpha_{k,\ell}^{(t)}\, p_\ell^{(t)}.
    \label{eq:mixture-update}
\end{equation}
By the MergeMix approximation, $\alpha_k^{(t)}$ identifies a point in
mixture space that sharpens domain $k$ while preserving other
objectives, and Eq.~\eqref{eq:mixture-update} uses this point as the
next-round training mixture for branch $k$. Composing these updates
across rounds carries information from each round's merge evaluations
into the next-round mixture, so $p_k^{(t)}$ reflects the full search
trajectory rather than the latest round alone.

\subsection{Multi-Objective Bayesian Optimization over the Merge Simplex}
\label{sec:mtm:bo}
The candidate set $\mathcal{X}_t$ in Eqs.~\eqref{eq:global-ratio}
and~\eqref{eq:specialist-ratio} is constructed by MOBO over the merge simplex. Each candidate
$\alpha\in\Delta^{K-1}$ defines a merged model
$\mathcal{M}^{(t)}(\alpha)$ and a normalized score vector
\begin{equation}
    \mathbf{F}_t(\alpha)
    =
    \bigl(F_t^{(1)}(\alpha),\ldots,F_t^{(K)}(\alpha)\bigr),
    \qquad
    F_t^{(j)}(\alpha)
    =
    f_j\!\left(\mathcal{M}^{(t)}(\alpha)\right).
\end{equation}
A dense grid over $\Delta^{K-1}$ becomes expensive as $K$ grows, while a
coarse grid can miss useful interior trade-offs. We therefore use MOBO to
adaptively propose merge weights that expand the Pareto frontier of
$\mathbf{F}_t(\alpha)$.

Each round is initialized with a small anchor set consisting of the
simplex vertices $\{e_k\}_{k=1}^{K}$, the uniform weight
$\tfrac{1}{K}\mathbf{1}$, and Dirichlet samples. Starting from these
observations, BO proposes additional batches of merge weights under the
simplex constraint $\alpha\in\Delta^{K-1}$. We use qNEHVI~\citep{daulton2021parallel} as the
acquisition function, since it handles noisy batched evaluations and
directly targets hypervolume improvement over multiple objectives.

MOBO is used only to construct the Pareto-aware candidate set
$\mathcal{X}_t$, not to directly choose the final merge. Once
$\mathcal{X}_t$ is obtained, MTM applies the scalarized selection rules in
Eqs.~\eqref{eq:global-ratio} and~\eqref{eq:specialist-ratio}: the averaged
score promotes the next reference model, while constrained
objective-specific scores update the branch mixtures. Thus, one batch of
merge evaluations per round supports both model promotion and all
branch-specific mixture updates. Algorithm~\ref{alg:imm-bo} summarizes
the full procedure.


\section{Main Experiments}
\label{sec:experiments}

We evaluate MTM in two settings that reflect the failure modes identified in \Cref{sec:background}: a \textbf{multi-domain} setting that merges math, code, and instruction-following branches (\Cref{sec:exp_multi}), and a \textbf{hybrid-mode} setting that merges non-think and think branches (\Cref{sec:exp_hybrid}).

\subsection{Experimental Setup}
\label{sec:exp_setup}

\paragraph{Training.}
We use two open base models that differ in scale and pretraining recipe:
Qwen3-4B-Base~\citep{yang2025qwen3} and OLMo-3-1025-7B (referred to as
OLMo-7B)~\citep{olmo2025olmo}. 
For the multi-domain setting, we build math, code, and instruction-following data from Dolci-Instruct-SFT\footnote{\url{https://huggingface.co/datasets/allenai/Dolci-Instruct-SFT}}
by sampling 10K examples from each corresponding category. We generate
model-specific trajectories using Qwen3-4B-Instruct\footnote{\url{https://huggingface.co/Qwen/Qwen3-4B-Instruct-2507}}
for Qwen3-4B and OLMo-Instruct\footnote{\url{https://huggingface.co/allenai/Olmo-3-7B-Instruct}}
for OLMo-7B. For the hybrid-mode setting, we sample 10K non-think examples from
Dolci-Instruct-SFT and 10K think examples from
Dolci-Think-SFT-32B\footnote{\url{https://huggingface.co/datasets/allenai/Dolci-Think-SFT-32B}},
generating think trajectories with Qwen3-4B\footnote{\url{https://huggingface.co/Qwen/Qwen3-4B}}
in think mode and the OLMo3-7B-Think for the respective experiments. Mixture
datasets contain 10K examples sampled from the branch-specific datasets, matching
the total training size of each individual branch. See \Cref{sec:app:exp_details} for details.

\paragraph{Benchmarks.}
We evaluate along three axes that correspond to the target objectives.
For \emph{Math}, we report accuracy on GSM8K~\citep{cobbe2021training} and MATH-500~\citep{hendrycks2021measuring}. For \emph{Code}, we report
MBPP+~\citep{austin2021program} and LiveCodeBench (LCB)~\citep{jain2024livecodebench}; For \emph{IF (Instruction Following)}, we report
IFEval~\citep{zhou2023instruction}. In the
hybrid setting we additionally report the average response length (in tokens)
under both \texttt{no-think} and \texttt{think} prompts, which measures
whether the merged model preserves behavioral separation between the two
modes. See \Cref{sec:app:exp_details} for evaluation prompts and additional details.

\paragraph{Baselines.}
We compare MTM against several baselines constructed from the same set of single-objective experts. \textbf{(1) Linear Merge}~\citep{ilharco2023editing} linearly combines expert updates, with merge weights chosen by grid search over the simplex. \textbf{(2) MergeMix}~\citep{wang2026mergemix} uses the selected merge weights as data-mixture ratios and trains a single model on the resulting mixture, rather than merging at deployment time. \textbf{(3) TIES-Merge}~\citep{yadav2023ties} resolves interference by sparsifying updates, electing a shared sign, and averaging sign-consistent parameters. \textbf{(4) DARE}~\citep{yu2024language} randomly drops and rescales update entries, and we apply it as a preprocessing step before TIES. \textbf{(5) SCE}~\citep{wan2025fusechat} assigns matrix-wise coefficients based on update magnitudes to enable fine-grained fusion without hard sparsification.


\subsection{Multi-Domain}
\label{sec:exp_multi}

\begin{table*}[t]
\centering
\scriptsize
\caption{\textbf{Results for Qwen3-4B-Base and OLMo-7B under the SFT setting (3-eval pass@1 avg $\pm$ std).} \textbf{Bold} marks the best result on the Avg.\ column and \underline{underline} marks the second best.}
\setlength{\tabcolsep}{2pt}
\renewcommand{\arraystretch}{0.95}

\resizebox{\textwidth}{!}{%
\begin{tabular}{lcccccc@{\hspace{1.2em}}cccccc}
\toprule
\multicolumn{7}{c}{\textbf{Qwen3-4B-Base}} &
\multicolumn{6}{c}{\textbf{OLMo-7B}} \\
\cmidrule(r){1-7} \cmidrule(l){8-13}
Method & GSM8K & MATH & MBPP+ & LCB & IFEval & Avg. &
GSM8K & MATH & MBPP+ & LCB & IFEval & Avg. \\
\midrule

Math Expert
& 92.85\std{0.22} & 84.40\std{0.93} & 64.02\std{1.63} & 20.59\std{0.48} & 55.95\std{0.39} & 63.56\std{0.73}
& 89.84\std{0.65} & 79.20\std{1.39} & 54.32\std{0.55} & 18.58\std{0.46} & 38.92\std{1.24} & 56.17\std{0.86} \\

Code Expert
& 79.53\std{6.63} & 53.85\std{11.13} & 67.46\std{0.49} & 22.03\std{1.40} & 67.99\std{0.53} & 58.17\std{4.04}
& 63.46\std{3.72} & 66.20\std{0.80} & 63.67\std{0.15} & 16.01\std{0.51} & 50.51\std{1.29} & 51.97\std{1.29} \\

IF Expert
& 90.01\std{0.47} & 76.40\std{0.71} & 67.92\std{0.82} & 18.72\std{0.78} & 80.57\std{0.50} & 66.72\std{0.66}
& 67.65\std{2.10} & 52.07\std{2.91} & 63.67\std{0.81} & 14.61\std{1.11} & 76.06\std{1.51} & 54.81\std{1.69} \\

\midrule

Linear Merge
& 91.91\std{0.38} & 83.10\std{0.59} & 67.72\std{0.42} & 23.40\std{0.29} & 77.26\std{3.05} & 68.68\std{0.95}
& 83.83\std{1.30} & 75.00\std{0.53} & 64.55\std{0.00} & 14.61\std{0.64} & 61.62\std{2.10} & 59.92\std{0.91} \\

Ties Merge
& 92.04\std{0.50} & 83.07\std{0.50} & 68.08\std{1.46} & 23.35\std{1.91} & 78.30\std{1.18} & 68.97\std{1.11}
& 87.87\std{0.39} & 76.20\std{1.97} & 63.05\std{1.76} & 14.90\std{1.04} & 60.76\std{2.35} & 60.56\std{1.50} \\

DARE-Ties
& 91.61\std{0.38} & 81.67\std{2.00} & 64.20\std{1.76} & 19.53\std{0.71} & 76.77\std{1.42} & 66.76\std{1.25}
& 88.40\std{0.85} & 76.73\std{0.90} & 59.26\std{0.79} & 16.45\std{0.64} & 60.74\std{1.94} & 60.32\std{1.03} \\

SCE
& 91.43\std{0.99} & 83.67\std{2.12} & 70.99\std{0.93} & 22.03\std{1.38} & 74.29\std{0.38} & 68.48\std{1.16}
& 81.15\std{1.21} & 75.53\std{2.05} & 59.61\std{4.71} & 14.24\std{3.77} & 56.95\std{1.80} & 57.50\std{2.71} \\

MergeMix
& 88.64\std{0.87} & 80.35\std{1.69} & 67.46\std{0.82} & 21.81\std{0.64} & 80.65\std{0.65} & 67.78\std{0.93}
& 89.21\std{0.61} & 76.93\std{0.50} & 62.17\std{0.26} & 12.92\std{0.25} & 74.81\std{0.50} & 63.21\std{0.43} \\

\midrule

MTM-Single
& 90.88\std{0.90} & 82.45\std{1.02} & 69.11\std{0.80} & 22.69\std{0.72} & 82.87\std{0.62} & \underline{69.60}\std{0.81}
& 89.76\std{0.66} & 78.13\std{0.46} & 62.35\std{1.36} & 13.00\std{0.58} & 74.10\std{0.15} & \underline{63.47}\std{0.64} \\

MTM-Multi
& 92.32\std{0.93} & 82.40\std{0.80} & 68.78\std{0.75} & 21.04\std{0.62} & 84.16\std{0.59} & \textbf{69.74}\std{0.74}
& 90.32\std{0.83} & 78.27\std{1.40} & 64.64\std{1.19} & 13.36\std{0.64} & 73.89\std{0.78} & \textbf{64.10}\std{0.97} \\

\bottomrule
\end{tabular}%
}
\label{tab:qwen_olmo_sft_3eval_horizontal}
\end{table*}
\Cref{tab:qwen_olmo_sft_3eval} shows that MTM-Multi achieves the
highest average score on both Qwen3-4B-Base and OLMo-7B, outperforming
all merging-only baselines. Across both models, MTM is the only method
that keeps math, code, and instruction-following performance
simultaneously close to the best single-objective experts, rather than
recovering one axis at the cost of another. MTM also consistently improves over MergeMix, which uses merge-based
search only to select a single global data mixture. This indicates that
the gains come from combining \emph{branch-specific} mixtures with
\emph{deployment-time merging}, not from mixture selection alone. The
gap is most pronounced on IF metrics, where MTM can
retain an IF-leaning branch and recombine it with other branches at merge
time.

\paragraph{Iterative refinement matters.}
MTM-Multi consistently outperforms MTM-Single, with the largest gains on
interference-sensitive metrics such as IFEval. This supports
the role of the iterative procedure in \Cref{sec:method}: each round
realigns branches around the promoted shared reference model and adapts
the next branch mixtures using specialist ratios.

\begin{figure*}[t]
  \centering

  \begin{minipage}[c]{0.35\textwidth}
    \centering
    \captionof{table}{\textbf{Relative epoch analysis.} Cost denotes total training budget; Path denotes the effective optimization path length of the returned model. Merge-only includes Linear Merge, TIES Merge, and DARE-TIES. MTM-Multi reports Qwen3-4B-Base/OLMo-7B.}
    \label{tab:epoch_analysis}

    \scriptsize
    \setlength{\tabcolsep}{4pt}
    \renewcommand{\arraystretch}{1.0}
    \resizebox{\textwidth}{!}{
      \begin{tabular}{lcc}
        \toprule
        Method family & Cost & Path \\
        \midrule
        Expert SFT & $1.0$ & $1.0$ \\
        Merge-only & $1.0$ & $1.0$ \\
        MergeMix / MTM-Single & $2.0$ & $1.0$ \\
        MTM-Multi & $\mathbf{0.70/0.75}$ & $\mathbf{0.70/0.75}$ \\
        \bottomrule
      \end{tabular}
    }
  \end{minipage}
  \hfill
  \begin{minipage}[c]{0.6\textwidth}
    \centering
    \includegraphics[width=\textwidth]{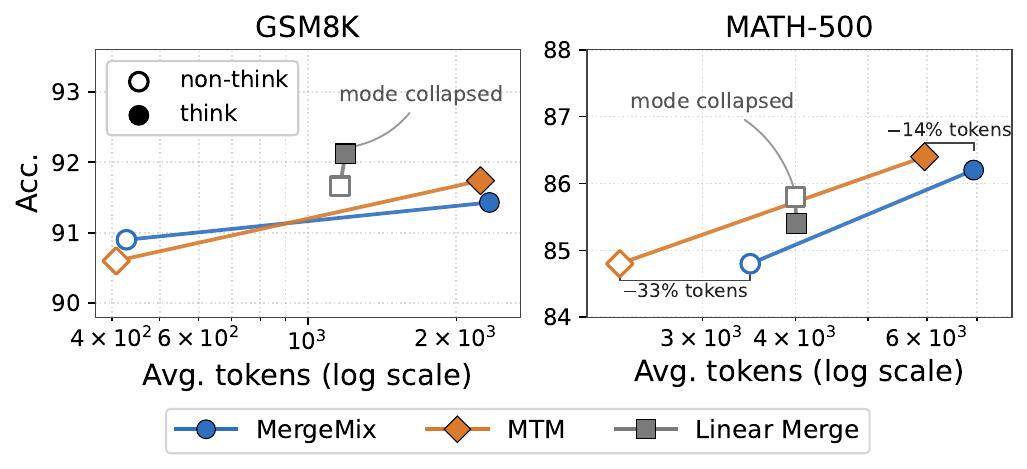}
    \captionof{figure}{\textbf{Length-accuracy frontier under \texttt{non-think}/\texttt{think} modes on OLMo-7B.} Open/filled markers share a model; segment length = mode separation, Linear Merge collapses both modes.}
    \label{fig:mode_exp}
  \end{minipage}
\end{figure*}

\paragraph{Training cost.}
\Cref{tab:epoch_analysis} compares the relative training budget and
effective optimization path length of each method. MTM introduces
additional cost by training mixture-biased branches, but these branches
are independent within each round and can be trained in parallel.
Moreover, MTM-Multi reuses the initial experts as the first stage of the
iterative procedure rather than discarding them. As a result, its
wall-clock cost remains comparable to a single-trajectory mixture
training run with the same number of stages, while preserving the
deployment-time flexibility of weight-space merging.


\subsection{Hybrid Mode}
\label{sec:exp_hybrid}

We now turn to the harder hybrid setting in which the two branches encode 
\emph{behaviors} rather than domains: a \texttt{no-think} branch trained to 
produce short, direct answers, and a \texttt{think} branch trained to 
produce long CoT reasoning. As shown in 
\Cref{sec:background}, naively merging these two branches collapses the two 
modes onto a single behavior, motivating mixture training and ultimately 
MTM as remedies. We probe this on OLMo-3-7B by comparing Linear Merge, 
Mixture Training, and MTM. Each method is evaluated on the same unified 
model under both \texttt{no-think} and \texttt{think} prompts, yielding 
two (token, accuracy) points per method (\Cref{fig:mode_exp}). The 
horizontal extent between the two points measures how strongly the modes 
diverge in generation length, while their vertical positions reflect 
accuracy: a short segment means the two modes have collapsed into 
near-identical behavior, whereas a longer segment positioned further to 
the upper-left means modes remain distinct while overall using fewer 
tokens at comparable accuracy.

\paragraph{MTM preserves mode separation and shifts the frontier.}
Reading \Cref{fig:mode_exp} under this convention, Linear Merge (gray) 
produces a near-degenerate segment, indicating that the two modes have 
collapsed into a single behavior. Mixture Training (blue) yields a 
clearly extended segment, recovering the separation between short 
\texttt{no-think} and long \texttt{think} generations. MTM (orange) 
preserves this separation and additionally shifts the segment toward 
the upper-left, using fewer tokens at \texttt{no-think} while matching 
or exceeding Mixture Training accuracy at \texttt{think}. Together, 
these results confirm that mode separation must be built in through 
mixture training rather than merging hard experts, and that MTM further 
exploits branch-level mixture control to push the hybrid model onto a 
more favorable length/accuracy trade-off than mixture training alone 
can reach.
\section{Additional Experiments}

\begin{figure*}[t]
  \centering

  \begin{minipage}[c]{0.5\textwidth}
    \centering
    \includegraphics[width=1.0\textwidth]{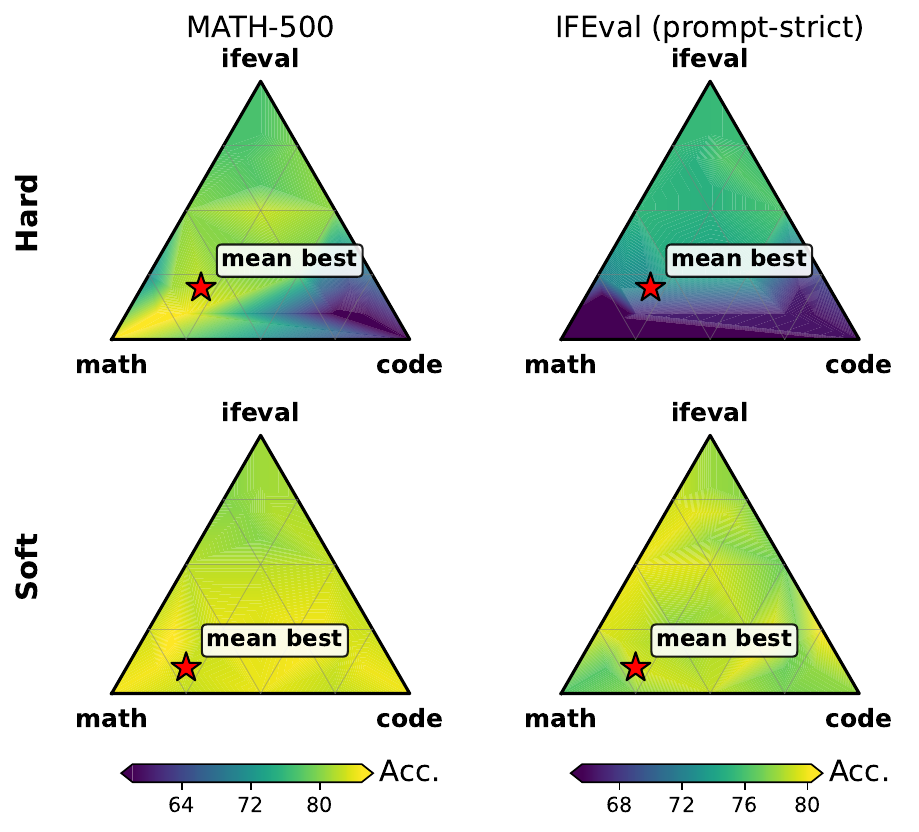}
    \captionof{figure}{\textbf{Merge-simplex accuracy on Qwen3-4B-Base.} Ternary heatmaps over $\alpha \in \Delta^2$ for \textbf{Hard} vs.\textbf{Soft} branches on MATH-500 and IFEval. \textbf{Soft} branches yield a flatter, higher-valued interior.}

    \label{fig:acc_surface_3way}
  \end{minipage}
  \hfill
  \begin{minipage}[c]{0.46\textwidth}
    \centering
    \captionof{table}{\textbf{Recipe transfer across architectures.}
MTM recipes (branch mixtures $\{p_k\}$ and merge coefficients
$\alpha_\star$) selected on Qwen3-4B-Base are reused on Qwen2.5-0.5B
without re-running the search. For a controlled comparison, all methods
are trained for the same fixed budget of 3 epochs. Merge (Uniform Merge)
and Mixture (Uniform Mixture) are size-matched baselines that ignore the
source recipe.}
    \resizebox{\textwidth}{!}{
    \begin{tabular}{lcccc}
      \toprule
        Method & MATH-500 & MBPP+ & IFEval & Avg. \\
        \midrule
        Merge & 37.00 & 27.78 & 37.65 & 34.14 \\
        Mixture & 32.40 & 24.07 & 43.63 & 33.37 \\
        MTM & 32.00 & 34.66 & 45.93 & \textbf{37.53} \\
      \bottomrule
    \end{tabular}
    }
    \label{tab:qwen25_transfer}
  \end{minipage}

\end{figure*}


\paragraph{Merge-simplex landscape.}
To visualize how mixture training reshapes the weight-space geometry exploited by MTM, we evaluate $\mathcal{M}^{(t)}(\alpha)$ on a dense grid of $\alpha \in \Delta^2$ for the three-branch (math, code, IF) setting on Qwen3-4B-Base, comparing \textbf{Hard} branches (single-domain SFT) against \textbf{Soft} branches (mixture-trained). Figure~\ref{fig:acc_surface_3way} shows that the Hard branches concentrate accuracy near the corresponding vertex and decay sharply along interior rays, consistent with independently drifted experts lying outside any shared high-accuracy basin. Soft branches yield a flatter, higher-valued interior, so the selectors $\alpha_\star^{(t)}$ in Eq.~\eqref{eq:promote-base} and $\alpha_k^{(t)}$ in Eq.~\eqref{eq:specialist-ratio} admit non-degenerate interior solutions that satisfy the cross-objective constraints $\tau_j^{(t)}$. This is direct geometric evidence that mergeability is a property of the branches, not the merger.

\paragraph{Cross-architecture recipe transfer.}
We next ask whether the recipe found by MTM is tied to the specific source model,
or whether it captures a more general structure of the multi-objective search
space. To test this, we take the branch mixtures $\{p_k\}$ and merge coefficients
$\alpha_\star$ selected on Qwen3-4B-Base and apply them directly to
Qwen2.5-0.5B, without any additional search or tuning. As shown in
\Cref{tab:qwen25_transfer}, MTM Transfer achieves a higher average score than
both size-matched baselines. This suggests that the learned recipe can serve as
a strong starting point for other architectures and model scales.
\section{Conclusion}

We study multi-objective language models built through weight-space merging. Merging single-objective experts often produces sharp cross-domain trade-offs or mode collapse in \texttt{think}/\texttt{non-think} hybrids. We attribute these failures to weight-space incompatibility: independently trained experts drift away from a shared high-performing
basin. We propose \textbf{Mixture-Trained Merging} (MTM), which trains each branch on an objective-biased mixture so that task vectors remain compatible. MTM uses merged-model evaluations as a proxy for data-mixture training and iterates between MOBO based mixture search, branch specialization, and merging. Across Qwen3-4B-Base and OLMo-7B, MTM-Multi achieves the best average performance, keeps all objectives close to their single-domain experts, and preserves \texttt{think}/\texttt{non-think} separation. The flatter,
higher-valued simplex induced by mixture-trained branches supports our main conclusion: \emph{mergeability is determined by the branches, not only by the merging rule}. Our study is limited to SFT-stage merging; extending MTM to larger objective sets and post-RL training remains an important direction for future work.

\clearpage

\begin{ack}
Use unnumbered first level headings for the acknowledgments. All acknowledgments
go at the end of the paper before the list of references. Moreover, you are required to declare
funding (financial activities supporting the submitted work) and competing interests (related financial activities outside the submitted work).
More information about this disclosure can be found at: \url{https://neurips.cc/Conferences/2026/PaperInformation/FundingDisclosure}.

Do {\bf not} include this section in the anonymized submission, only in the final paper. You can use the \texttt{ack} environment provided in the style file to automatically hide this section in the anonymized submission.
\end{ack}

\bibliographystyle{unsrt}
{
\small
\bibliography{references}
}

\clearpage


\appendix
\section{Experimental Details}
\label{sec:app:exp_details}

\subsection{Detail for Qwen3-4B-Base}

We fine-tune Qwen3-4B-Base\footnote{\url{https://huggingface.co/Qwen/Qwen3-4B-Base}}
on trajectories generated by Qwen3-4B-Instruct-2507\footnote{\url{https://huggingface.co/Qwen/Qwen3-4B-Instruct-2507}}
in \textsc{nothink} mode. Prompts are drawn from Dolci-Instruct-SFT\footnote{\url{https://huggingface.co/datasets/allenai/Dolci-Instruct-SFT}}
across three domains: math, coding, and precise-IF. We construct a separate corpus for each domain. Trajectories are generated using vLLM\footnote{\url{https://github.com/vllm-project/vllm}}
with nucleus sampling ($T{=}0.6$, top-$p{=}0.95$, top-$k{=}20$) and a maximum of $8{,}192$ new tokens.
Samples exceeding $15{,}000$ total tokens are filtered out, yielding $10{,}000$ examples per training run.
We use the same chat template for both training and inference.

For mixture-data SFT (used in the BO branch retrains; \S\ref{appendix:bo}),
we sample domain proportions $(\lambda_\mathrm{math}, \lambda_\mathrm{code}, \lambda_\mathrm{if}) \in \Delta^2$
from the per-domain corpora.

Training is performed on $4{\times}$ H200 GPUs using DeepSpeed ZeRO-2, bf16 weights, and TF32 matmul.
The loss is computed only over assistant tokens.

Instead of a fixed epoch budget, we apply plateau-based early stopping.
Training stops when the EMA ($\beta{=}0.9$) of validation token-level accuracy
(512 samples, evaluated every 20 steps) fails to improve by $\tau{=}0.02$
for three consecutive evaluations after a 100-step warm-up.

For BO branch retraining, we additionally impose a maximum step budget
with geometric decay: $T_c = T_1 r^{c-1}$, where $r{=}0.5$.

\begin{table}[h]
\centering
\caption{SFT hyperparameters for Qwen3-4B-Base.}
\vspace{0.5em}
\small
\begin{tabular}{ll}
\toprule
Optimizer / LR & AdamW (fused), $2{\times}10^{-5}$, constant w/ $0.03$ warmup \\
Weight decay / clip & $0.01$ / $1.0$ \\
Effective batch size & $128$ ($2 \times 16 \times 4$ GPUs) \\
Max sequence length & $10{,}000$ \\
Precision & bf16 weights, tf32 matmul, grad ckpt \\
Distributed & DeepSpeed ZeRO-2 \\
Plateau stop ($\tau,K,\beta$) & $0.02$, $3$, $0.9$ (val 512, every 20 steps) \\
Seed & $42$ \\
\bottomrule
\end{tabular}
\label{tab:sft-hparams-qwen}
\end{table}

\subsection{Detail for OLMo-7B}

We fine-tune OLMo-3-1025-7B\footnote{\url{https://huggingface.co/allenai/Olmo-3-1025-7B}}
under two settings: (1) multi-domain and (2) hybrid-mode (think / nothink).

In the multi-domain setting, we generate math, coding, and precise-IF trajectories
using Olmo-3-7B-Instruct\footnote{\url{https://huggingface.co/allenai/Olmo-3-7B-Instruct}}
with prompts from Dolci-Instruct-SFT. Each domain forms a separate corpus.

In the hybrid-mode setting, we additionally include \textsc{think} trajectories
generated by Olmo-3-7B-Think\footnote{\url{https://huggingface.co/allenai/Olmo-3-7B-Think}}
using prompts from Dolci-Think-SFT-32B\footnote{\url{https://huggingface.co/datasets/allenai/Dolci-Think-SFT-32B}}.
These trajectories prepend a $\langle\texttt{think}\rangle \dots \langle/\texttt{think}\rangle$
reasoning block before the final answer, while \textsc{nothink} trajectories directly output the answer.

The sampling configuration matches that of the Qwen setup.

Training is conducted on $4{\times}$ H200 GPUs with DeepSpeed ZeRO-2, bf16 weights, and TF32 matmul.
The loss is computed over assistant tokens only, including the reasoning block in \textsc{think} mode.

Unlike Qwen, we use a fixed epoch budget without early stopping.
Hyperparameters are summarized in Table~\ref{tab:sft-hparams-olmo}.

\begin{table}[h]
\centering
\caption{SFT hyperparameters for OLMo-7B. Identical across
\textsc{think} and \textsc{nothink} runs; only the input corpus and
trajectory length cap differ ($16{,}384$ vs.\ $8{,}192$ new tokens).}
\vspace{0.5em}
\small
\begin{tabular}{ll}
\toprule
Optimizer / LR & AdamW (fused), $2{\times}10^{-5}$, constant w/ $0.03$ warmup \\
Weight decay / clip & $0.01$ / $1.0$ \\
Effective batch size & $64$ ($1 \times 16 \times 4$ GPUs) \\
Max sequence length & $10{,}000$ \\
Precision & bf16 weights, tf32 matmul, grad ckpt \\
Distributed & DeepSpeed ZeRO-2 \\
Early stopping & disabled (fixed epoch budget) \\
Seed & $42$ \\
\bottomrule
\end{tabular}
\label{tab:sft-hparams-olmo}
\end{table}

\subsection{Bayesian Optimization Details}
\label{appendix:bo}

We use Bayesian optimization to search over merge coefficients on the
simplex. At each MTM iteration, the initial design includes the simplex
vertices, a central or previously best merge point, and a small number
of space-filling random candidates. Each candidate corresponds to a
merged checkpoint, which is evaluated on all objectives.

Each objective is modeled with an independent Gaussian process surrogate
using a Mat\'ern-5/2 kernel with automatic relevance determination and
outcome standardization. The surrogates are fit by maximizing the
marginal likelihood and refit after each new observation. We use qNEHVI
as the acquisition function, optimized sequentially with batch size
$q=1$. The reference point is defined relative to the single-objective
expert scores, encouraging hypervolume improvement only for candidates
that approach the expert frontier while maintaining balanced
performance across objectives.

After exhausting the BO budget, we select a global merge coefficient
from the observed candidates using a balanced multi-objective score that
prioritizes the weakest objective and avoids near-vertex solutions. We
also select specialist merge coefficients for the next MTM iteration by
requiring the target objective to remain close to its expert score while
constraining degradation on the other objectives. These coefficients are
then used for base promotion and mixture propagation in the next
iteration.

\subsection{Evaluation Settings}
\label{sec:app:eval_settings}

\begin{table}[h]
\centering
\small
\caption{Evaluation settings for each benchmark. $n$ denotes the
number of generations per prompt. All benchmarks use one greedy
generation and three sampled generations with $T{=}0.7$ and
top-$p{=}0.95$ across different random seeds.}
\vspace{0.5em}
\setlength{\tabcolsep}{5pt}
\begin{tabular}{lccc}
\toprule
Benchmark & \# Samples & $n$ & Metric \\
\midrule
IFEval\footnote{\url{https://huggingface.co/datasets/google/IFEval}}
  & 541   & 4 & prompt/inst.\ acc.\ (strict, avg.) \\
MBPP+\footnote{\url{https://github.com/evalplus/evalplus}}
  & 378   & 4 & pass@1 (avg.) \\
MATH-500\footnote{\url{https://huggingface.co/datasets/HuggingFaceH4/MATH-500}}
  & 500   & 4 & pass@1 (avg.) \\
GSM8K\footnote{\url{https://huggingface.co/datasets/openai/gsm8k} (test split)}
  & 1{,}319 & 4 & exact match (avg.) \\
LiveCodeBench\footnote{\url{https://livecodebench.github.io/}}
  & 268 (v5, 2024-08\,--\,2025-02) & 4 & pass@1 (avg.) \\
\bottomrule
\end{tabular}
\label{tab:eval-settings}
\end{table}

We evaluate all models using the NeMo-Skills
framework\footnote{\url{https://github.com/NVIDIA/NeMo-Skills}},
serving checkpoints with
vLLM\footnote{\url{https://github.com/vllm-project/vllm}} on
$16{\times}$ NVIDIA H200 80GB GPUs. We use tensor parallelism
$\mathrm{TP}\in\{2,4\}$ and bfloat16 weights. The maximum generation
length is set between 8,196 and 30,000 tokens depending on the
benchmark. Generations are scored using the official evaluators
bundled with NeMo-Skills. For each prompt, we generate four responses:
one greedy response with $T{=}0$ and three sampled responses using
$T{=}0.7$, top-$p{=}0.95$, and different random seeds. We report the
average score across these four generations. Dataset sizes and metrics
are summarized in \Cref{tab:eval-settings}.
\section{Algorithm}

\begin{algorithm}[t]
\caption{Iterative Mixture--Merge with Bayesian Optimization}
\label{alg:imm-bo}
\begin{algorithmic}[1]
\REQUIRE Base model $M_0$; corpora $\{D_k\}_{k=1}^{K}$; iterations $T$; BO budget $B$; thresholds $\{\tau_j^{(t)}\}$
\STATE Initialize $p_k^{(1)}\gets e_k$ for all $k$
\FOR{$t=1,\ldots,T$}
    \STATE Train experts:
    $E_k^{(t)}\gets
    \textsc{FineTune}\!\left(M_{t-1},\textsc{Mix}(D_{1:K};p_k^{(t)})\right)$
    for all $k$
    \STATE Initialize candidate set $\mathcal{X}_t$ with vertices $\{e_k\}$, the uniform weight, and random Dirichlet samples
    \FOR{$b=1,\ldots,B$}
        \STATE $\alpha_b\gets\textsc{BO-Propose}(\mathcal{X}_t)$
        \STATE Evaluate
        $f_k^{(t)}(\alpha_b)\gets
        \textsc{Eval}_k\!\left(\textsc{Merge}(E_{1:K}^{(t)};\alpha_b)\right)$
        for all $k$
        \STATE $\mathcal{X}_t\gets\mathcal{X}_t\cup\{\alpha_b\}$
    \ENDFOR
    \STATE Select global ratio
    $r_\star^{(t)}\gets
    \arg\max_{\alpha\in\mathcal{X}_t}
    \tfrac{1}{K}\sum_{j=1}^{K}f_j^{(t)}(\alpha)$
    \STATE Select each specialist ratio $r_k^{(t)}$ by constrained target-domain performance
    \STATE Promote base:
    $M_t\gets\textsc{Merge}(E_{1:K}^{(t)};r_\star^{(t)})$
    \STATE Update mixtures:
    $p_k^{(t+1)}\gets\sum_{j=1}^{K} r_{k,j}^{(t)}\, p_j^{(t)}$
    for all $k$
\ENDFOR
\RETURN $M_T$
\end{algorithmic}
\end{algorithm}

Algorithm~\ref{alg:imm-bo} provides the full MTM iteration, including
branch training, BO-based merge-ratio search, base promotion, and
mixture updates.



\end{document}